\documentclass{article} 
\usepackage{iclr2026_conference,times}

\usepackage{amsmath,amsfonts,bm}

\def\eqref#1{equation~\ref{#1}}

\def\1{\bm{1}}

\def\vc{{\bm{c}}}

\def\vv{{\bm{v}}}

\def\vx{{\bm{x}}}

\DeclareMathAlphabet{\mathsfit}{\encodingdefault}{\sfdefault}{m}{sl}
\SetMathAlphabet{\mathsfit}{bold}{\encodingdefault}{\sfdefault}{bx}{n}

\usepackage{tabularx}
\usepackage{multirow}
\usepackage{booktabs}
\usepackage[table]{xcolor}
\usepackage[dvipsnames]{xcolor}
\usepackage{graphicx}
\usepackage{adjustbox}
\usepackage{booktabs}
\usepackage{subcaption}

\usepackage{hyperref}
\usepackage{url}
\usepackage{graphicx}
\usepackage{enumitem}

\usepackage{wrapfig}
\usepackage{xcolor}
\usepackage{algorithm}
\usepackage{algpseudocode}
\algrenewcommand\algorithmiccomment[1]{\hfill\small\textcolor{gray}{\textit{// #1}}}

\definecolor{CQColor}{rgb}{0.0,0.0,1.0} 

\definecolor{TSColor}{rgb}{0.5,0.0,0.8} 

\definecolor{CQRColor}{rgb}{1.0,0.0,0.0} 

\definecolor{mycolor_blue}{HTML}{E7EFFA}
\definecolor{mycolor_green}{HTML}{E6F8E0}
\definecolor{mycolor_gray}{HTML}{ECECEC}
\definecolor{pearDark}{HTML}{2980B9}
\definecolor{citecolor}{HTML}{2980b9}
\definecolor{linkcolor}{HTML}{c0392b}

\definecolor{cL}{HTML}{5B8EC2}  
\definecolor{cR}{HTML}{6B9E5A}  
\title{%
  {\color{cL}O}%
  {\color{cL!83!cR}m}%
  {\color{cL!67!cR}n}%
  {\color{cL!50!cR}i}%
  {\color{cL!33!cR}N}%
  {\color{cL!17!cR}F}%
  {\color{cR}T}%
  : Modality-wise Omni Diffusion Reinforcement for Joint Audio-Video Generation%
}
\author{Guohui Zhang\textsuperscript{1},
Xiaoxiao Ma\textsuperscript{1},  
Jie Huang\textsuperscript{1},  
Hang Xu\textsuperscript{1},  
Hu Yu\textsuperscript{1},
Siming Fu\textsuperscript{3},
Yuming Li\textsuperscript{2}, \\
\textbf{Zeyue Xue}\textsuperscript{3},  
\textbf{Lin Song}\textsuperscript{3\ddag},
\textbf{Haoyang Huang}\textsuperscript{3},
\textbf{Nan Duan}\textsuperscript{3},
\textbf{Feng Zhao}\textsuperscript{1\dag} \\
\textsuperscript{1}University of Science and Technology of China,
\textsuperscript{2}Peking University,
\textsuperscript{3}JD Explore Academy
}

\iclrfinalcopy
\begin{document}

{
  \renewcommand{\thefootnote}{\fnsymbol{footnote}}
  \footnotetext{\textsuperscript{\dag}Corresponding author}
  \footnotetext{\textsuperscript{\ddag}Project leader}
}

\maketitle

\begin{figure}[h]
\centering
    \vspace{-6mm}
   \includegraphics[width=0.98\textwidth]{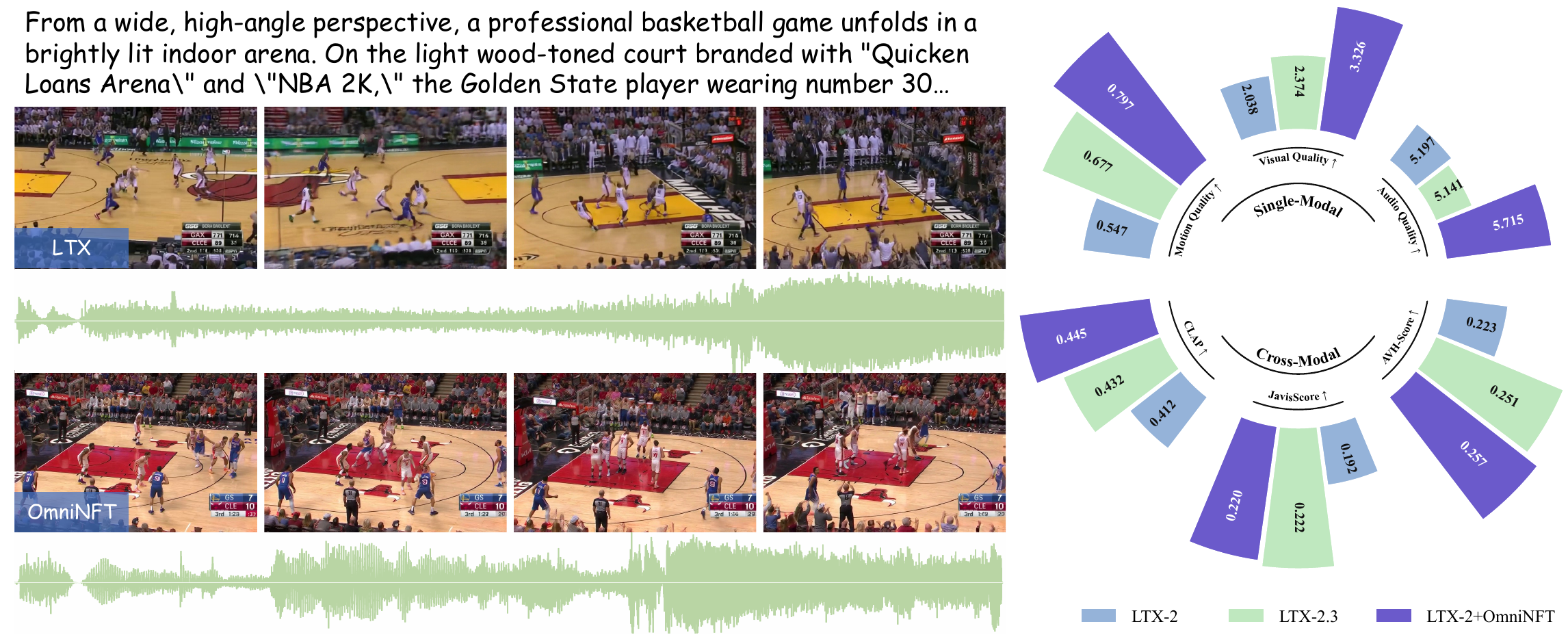}
    \caption{
  OmniNFT consistently improves the performance of LTX-2 in audio and visual quality, motion quality, cross-modal alignment, and audio–video synchronization. 
}
\label{fig:model_teaser}
\end{figure}%

\begin{abstract}
Recent advances in joint audio-video generation have been remarkable, yet real-world applications demand strong per-modality fidelity, cross-modal alignment, and fine-grained synchronization.
Reinforcement Learning (RL) offers a promising paradigm, but its extension to \textit{multi-objective} and \textit{multi-modal} joint audio-video generation remains unexplored. Notably, our in-depth analysis first reveals that the primary obstacles to applying RL in this stem from:
\textbf{(i)} multi-objective advantages inconsistency, where the advantages of multimodal outputs are not always consistent within a group; 
\textbf{(ii)} multi-modal gradients imbalance, where video-branch gradients leak into shallow audio layers responsible for intra-modal generation;
\textbf{(iii)} uniform credit assignment, where fine-grained cross-modal alignment regions fail to get efficient exploration.
These shortcomings suggest that vanilla RL fine-tuning strategy with a single global advantage often leads to suboptimal results.
To address these challenges, we propose \textbf{OmniNFT}, a novel modality-aware online diffusion RL framework with three key innovations: 
\textbf{(1)} Modality-wise advantage routing, which routes independent per-reward advantages to their respective modality generation branches.
\textbf{(2)} Layer-wise gradient surgery, which selectively detaches video-branch gradients on shallow audio layers while retaining those for cross-modal interaction layers.
\textbf{(3)} Region-wise loss reweighting, which modulates policy optimization toward critical regions related to audio-video synchronization and fine-grained alignment.
Extensive experiments on JavisBench and VBench with the LTX-2 backbone demonstrate that OmniNFT achieve comprehensive improvements in audio and video perceptual quality, cross-modal alignment, and audio-video synchronization. 

\paragraph{Project Page:} \url{https://zghhui.github.io/OmniNFT/}
\end{abstract}
\section{Introduction}
\label{sec:intro}

Recent years have witnessed significant advancements in joint audio-video generation~\citep{low2025ovi,liu2026javisdit++,seedance2025seedance}. However, achieving genuine practical utility in real-world applications demands a combination of high per-modality fidelity, robust cross-modal semantic consistency, and fine-grained audio-video synchronization. How to effectively align audio-video generative models with these multifaceted requirements remains an open and pressing challenge.

Despite rapid progress, current joint audio-video generative models~\citep{wang2025universe,hacohen2026ltx,liu2026javisdit++} still struggle to simultaneously satisfy these multifaceted objectives. Meanwhile, Reinforcement Learning with Verifiable Rewards (RLVR), particularly Group Relative Policy Optimization (GRPO)~\citep{guo2025deepseek}, has recently emerged as a powerful post-training paradigm for generative models~\citep{he2025tempflow,li2025mixgrpo}, owing to its ability to optimize complex and highly subjective objectives~\citep{liu2026gdpo}. The previous research~\citep{xue2025dancegrpo,zheng2025diffusionnft} has successfully leveraged RLVR to enhance text-to-image/video generation quality and semantic alignment. These successes motivate a key question: \textbf{\textit{can RLVR be effectively extended to joint audio-video generation to optimize fidelity, alignment, and AV synchronization?}}

\begin{figure}[t]
    \centering
    \includegraphics[width=\linewidth]{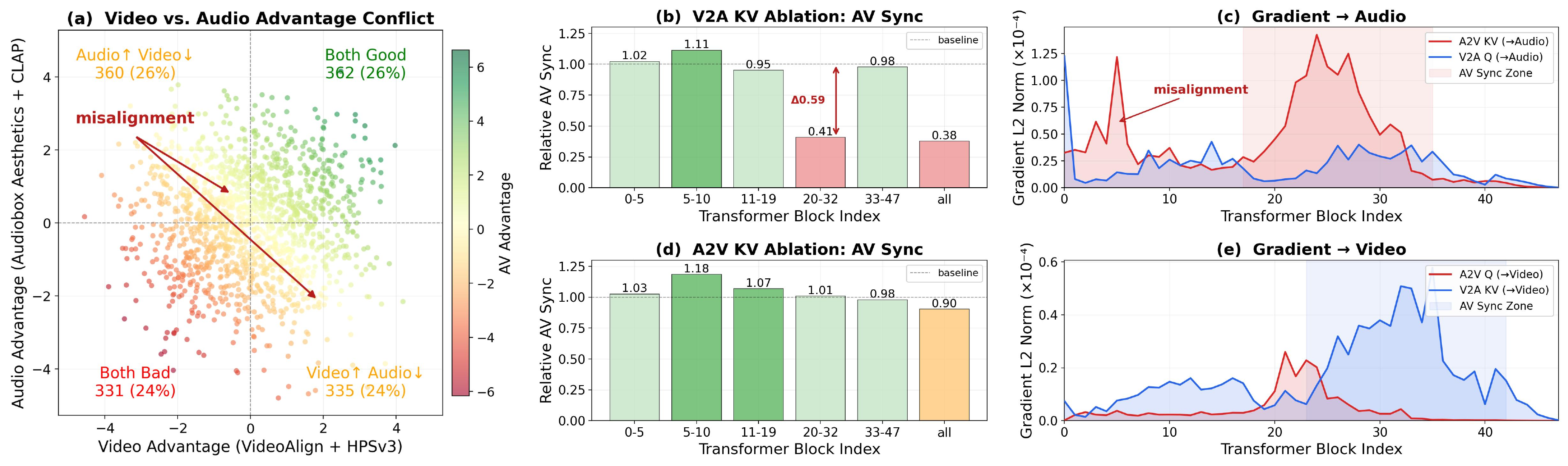}
    \caption{\textbf{Advantage inconsistency and asymmetric audio-video interaction.}
    (a) Video and audio advantages are weakly correlated: roughly half of the samples receive \emph{opposing} rewards across the two modalities.
    (b) Blocking the V2A $\mathrm{KV}$ in mid-layers collapses AV synchronization to $0.41\times$ baseline, whereas (d) the symmetric A2V ablation causes a mild degradation when applied to the later blocks.
    (c,e) Layer-wise gradient norms of cross-attention show that audio-video interaction is concentrated in the \emph{middle} and \emph{later} transformer blocks (AV Sync Zone). On the audio branch, gradients from video $\mathrm{KV}$ (A2V) disturb the update direction of the audio shallow blocks in (c).}
    \label{fig:av_conflict}
\end{figure}

However, the performance of directly applying vanilla RLVR to joint audio-video generation remains suboptimal. Our in-depth analysis reveals that the primary obstacles stem from three types of optimization mismatch:
\textbf{(i)} \emph{multi-objective advantages inconsistency}: the different rewards for the video and audio components of a single multimodal output are often inconsistent, as illustrated in Fig.~\ref{fig:av_conflict}(a).
\textbf{(ii)} \emph{multi-modal gradients imbalance}: intra-modal generation and cross-modal interaction mainly concentrate in the shallow and deeper layers, respectively, see Fig.~\ref{fig:av_conflict}. However, gradients from the video branch tend to dominate the update direction of the shallow audio layers in the backward process.
\textbf{(iii)} \emph{uniform credit assignment}: AV synchronization and fine-grained alignment focus on certain critical regions, yet uniform updates ignore different contributions of these regions.

We address these challenges with~\textbf{OmniNFT} (Modality-wise \textbf{Omni} Diffusion \textbf{N}egative-aware \textbf{F}ine-\textbf{T}uning), a novel modality-aware diffusion RL framework for joint audio-video generation. OmniNFT addresses the above issues with three coordinated designs. \textbf{(1) Modality-wise advantage routing:} for advantage inconsistency, we compute an independent advantage for each reward and selectively route it according to its underlying modality. \textbf{(2) Layer-wise gradient surgery:} for gradients imbalance, we detach the part gradient from the video stream on the shallow layers of the audio model, while preserving the effective gradients responsible for audio–video interaction. \textbf{(3) Region-wise loss reweighting:} for credit assignment, we introduce a critical-region reweighting strategy that strengthens the optimization on critical regions. This fine-grained credit assignment, coupled with gradient surgery, effectively circumvents optimization mismatch across modalities.

We conduct extensive experiments on JavisBench~\citep{liu2025javisdit} and VBench~\citep{huang2024vbench} using LTX-2~\citep{hacohen2026ltx} as the backbone. OmniNFT achieves comprehensive improvements across audio and video perceptual quality, cross-modal alignment, and audio-video synchronization. Our contributions are summarized as follows:
\begin{itemize}
    \item We first visit RL for joint audio-video generation and identify its core optimization bottleneck, including advantage inconsistency, gradients imbalance, and uniform credit assignment, which together hinder effective joint improvement.
    \item Based on these bottlenecks, we propose OmniNFT that combines modality-wise advantage routing, layer-wise gradient surgery, and region-wise loss reweighting, effectively facilitating multi-objective and multi-modal optimization.
    \item Extensive experiments demonstrate that OmniNFT delivers consistent gains in perceptual quality, cross-modal alignment, and audio-video synchronization over strong baselines.
\end{itemize}

\section{Related Work}

\subsection{Text-to-Video Generation.}
Text-to-video generation has rapidly progressed from U-Net-based diffusion to large-scale Transformer-based diffusion. Early methods~\citep{blattmann2023stable,guo2023animatediff,wu2023tune,blattmann2023align} typically adapt image diffusion backbones with temporal modules to support video generation, while recent Diffusion Transformers (DiTs)~\citep{brooks2024video,hong2022cogvideo,kong2024hunyuanvideo} scale more effectively with data and model size, demonstrating substantial gains in visual fidelity, physical plausibility, and text alignment. Among open-source lines, Wan~\citep{wan2025wan} emphasizes efficient spatiotemporal attention with flow-matching training, and LTX-Video~\citep{hacohen2024ltx} further targets real-time inference by jointly optimizing a Video-VAE and a denoising Transformer. 

\subsection{Joint Audio-Video Generation.}
Building on strong text-to-video backbones, recent work increasingly focuses on generating temporally synchronized video and audio in a single pipeline. Veo3~\citep{veo3} demonstrates joint audio-video capabilities. UniVerse-1~\citep{wang2025universe} combines Wan 2.1~\citep{wan2025wan} for video generation and ACE-Step~\citep{gong2025ace} for audio generation, and introduces lightweight projection modules to align latent spaces across modalities. In parallel, the Javis series~\citep{liu2025javisdit,liu2026javisdit++} explores unified modeling for joint generation, whereas the LTX series~\citep{hacohen2026ltx} adopts an asymmetric dual-stream design with bidirectional cross-attention to handle modality imbalance. While these efforts have yielded promising results, there remains substantial room for improvement in generation quality and audio-visual synchronization.

\subsection{Reinforcement Learning for Visual Generation.}
 Inspired by the success of GRPO~\citep{guo2025deepseek} in large language models (LLMs), RLVR has emerged as a practical post-training strategy for improving visual quality and semantic alignment in visual generation~\citep{luo2025reinforcement,zhang2025maskfocus,zhang2025group,jiang2025t2i}. Several methods adapt reward-driven policy optimization to image and video generation tasks. For example, T2I-R1~\citep{jiang2025t2i} introduces dual-level chain-of-thought RL for autoregressive generation, while Flow-GRPO~\citep{liu2025flow} and Dance-GRPO~\citep{xue2025dancegrpo} extend flow matching with stochastic trajectories by reformulating Ordinary Differential Equation (ODE) as Stochastic Differential Equation (SDE) processes. DiffusionNFT~\citep{zheng2025diffusionnft} instead finetunes diffusion models in the forward process through an implicit policy-improvement direction. However, effective expansion of RL into multi-objective joint audio-visual generation remains under-explored.

\section{Preliminary}
\label{sec:Preliminary}

\textbf{Joint Audio-Video Flow Matching.}
In joint audio-video generation, an audio latent $x^a$ and a video latent $x^v$ are denoised in parallel under a shared flow matching~\citep{lipman2022flow} schedule. Each modality is independently perturbed by its own Gaussian prior $x_1^m \sim \mathcal{N}(0, I)$, while sharing a timestep $t \in [0, 1]$:
\begin{equation}
    x_t^m = (1 - t)\, x_0^m + t\, x_1^m, \quad m \in \{a, v\}.
\end{equation}
A dual-stream model $v_\theta = (v_\theta^a, v_\theta^v)$ jointly predicts the velocity of both modalities, where the two streams interact through cross-modal attention layers. During the sampling phase, both streams are integrated in parallel via a deterministic ODE solver:
\begin{equation}
    dx_t^m = v_\theta^m(x_t^a, x_t^v, t, c)\, dt, \quad m \in \{a, v\},
\end{equation}
where $c$ denotes the text condition.

\textbf{Diffusion Negative-aware Finetuning (DiffusionNFT)}.  In contrast to existing diffusion-based GRPO frameworks~\citep{liu2025flow,xue2025dancegrpo}, which typically necessitate a transition from deterministic ODEs to SDEs, DiffusionNFT~\citep{zheng2025diffusionnft} performs policy optimization directly on the forward diffusion process. The method leverages a reward $r(x_0, c)$ to determine positive and negative policies, thereby defining a contrastive loss. The model’s velocity predictor, $v_\theta$, is encouraged toward the high-reward policy and away from the low-reward one. The core policy optimization loss is defined as:
\begin{equation}
\begin{split}
    \mathcal{L}(\theta) = \mathbb{E}_{c, \pi^{old}(x_0|c), t} \Big[
    r \| v_\theta^+(x_t, c, t) &- v \|_2^2 + 
     (1 - r) \| v_\theta^-(x_t, c, t) - v \|_2^2 \Big], \label{eq:loss}
\end{split}
\end{equation}
The implicit positive and negative policies $v_\theta^+$ and $v_\theta^-$ are combinations of the old policy $v^{old}$ and the training policy $v_\theta$, weighted by a hyperparameter $\beta$:
\begin{equation}
    v_\theta^+(x_t, c, t) := (1 - \beta) v^{old}(x_t, c, t) + \beta v_\theta(x_t, c, t),
\end{equation}
\begin{equation}
    v_\theta^-(x_t, c, t) := (1 + \beta) v^{old}(x_t, c, t) - \beta v_\theta(x_t, c, t).
\end{equation}
The optimality probability $r \in [0, 1]$ is transformed from the unconstrained raw reward signal $r^{raw}$:

\begin{equation}
    r(x_0, c) := \frac{1}{2} + \frac{1}{2} \text{clip} \left[ A(x_0, c), -1, 1 \right], \quad A(x_0, c) = \frac{r^{raw}(x_0, c) - \mathbb{E}_{\pi^{old}(\cdot|c)} r^{raw}(x_0, c)}{Z_c}, \label{eq:opt_prob}
\end{equation}

where $A(x_0, c)$ denotes the group-wise normalized advantage of sample $x_0$ under prompt $c$, $Z_c > 0$ is a normalizing factor, which could take the form of a global reward $std$.

\section{Motivation}
\label{sec:motivation}
\subsection{Advantage Inconsistency between Audio and Video}

\begin{wrapfigure}{r}{0.4\textwidth} 
    \centering 
    \vspace{-\intextsep} 
    \includegraphics[width=\linewidth]{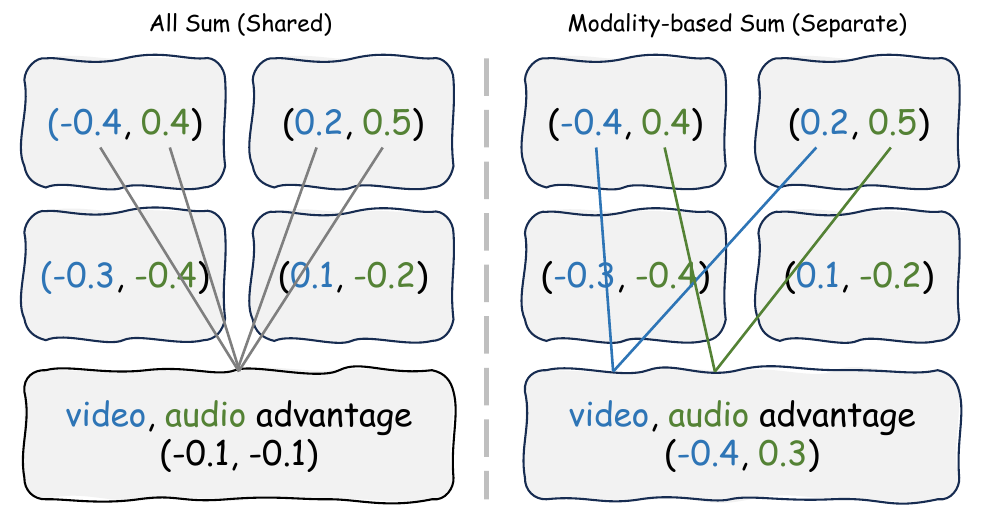}
    \caption{Advantage conflict between audio and video modality} 
    \label{fig:gdpo} 
    \vspace{-\intextsep} 
\end{wrapfigure}
In joint audio-video generation, the multimodal output needs to account for both video and audio quality simultaneously, yet the relationship between these two qualities has not been sufficiently analyzed. To this end, we analyze 1,400 generated samples (175 prompts, Group Size is 8) and calculate separate advantages for video and audio rewards. \textit{We observe that the rewards for video and audio of a single multimodal output are not often consistent}, as illustrated in Fig.~\ref{fig:av_conflict}(a). Specifically, high-quality videos are not necessarily accompanied by high-quality audio. On the contrary, nearly half of the samples exhibit advantage conflicts. We further enumerate four rollouts within a group, each represented as (video advantage, audio advantage), as shown in Fig.~\ref{fig:gdpo}. The All Sum (Shared) strategy collapses both modalities into a single scalar, yielding an indistinguishable and dilute advantage of ($-0.1, -0.1$). In contrast, the Modality-based Sum (Separate) strategy aggregates each modality independently, yielding ($-0.4, 0.3$), which faithfully reflects that video should be penalized while audio should be encouraged, enabling more informative learning.

\subsection{Gradients Imbalance across Different Modalities Branches}

The advantage inconsistency discussed above only reflects the issue in the output. We examine the internal information and gradients flow within the dual-stream model from both \textbf{forward} and \textbf{backward} perspectives, as illustrated in Fig.~\ref{fig:av_conflict}.

\textbf{Forward Analysis.}
We identify the functional roles of the Transformer blocks at different depths by selectively blocking the key-value (KV) information exchanged through audio-video cross-attention. As shown in Fig.~\ref{fig:av_conflict}(b) and (d), ablating the KV flow in shallow blocks (blocks 0-19) causes only marginal degradation in AV synchronization. In contrast, disrupting the KV flow in middle or deep blocks (blocks 20-32 for audio and 25-47 for video) leads to a substantial drop in AV alignment, with the audio branch exhibiting a relative decrease of up to $\Delta 0.59$. This indicates that \textit{shallow blocks are primarily responsible for intra-modal generation, while middle-to-deep blocks handle cross-modal audio-video interaction and alignment (AV-Sync Zone)}.

\textbf{Backward Analysis.}
Within the AV-Sync Zone, gradients flowing through the cross-modal interaction paths are dominant, indicating that the RL signal is correctly concentrated on the blocks truly responsible for cross-modal alignment. Meanwhile, the magnitudes of these interaction gradients gradually decay toward shallower layers. However, we observe an {anomalous gradient spike} in the shallow layers (Fig.~\ref{fig:av_conflict}(c)) on the audio branch, implying that a large portion of the RL gradient is erroneously injected into layers dedicated to intra-modal generation. We argue that this {gradient misalignment} disrupts the intra-modal audio generation process, resulting in sub-optimal results.

\subsection{V2A Cross-Attention as an Intrinsic Proxy for Critical Regions}

\begin{wrapfigure}{r}{0.4\textwidth} 
    \centering 
    \vspace{-\intextsep} 
    \includegraphics[width=\linewidth]{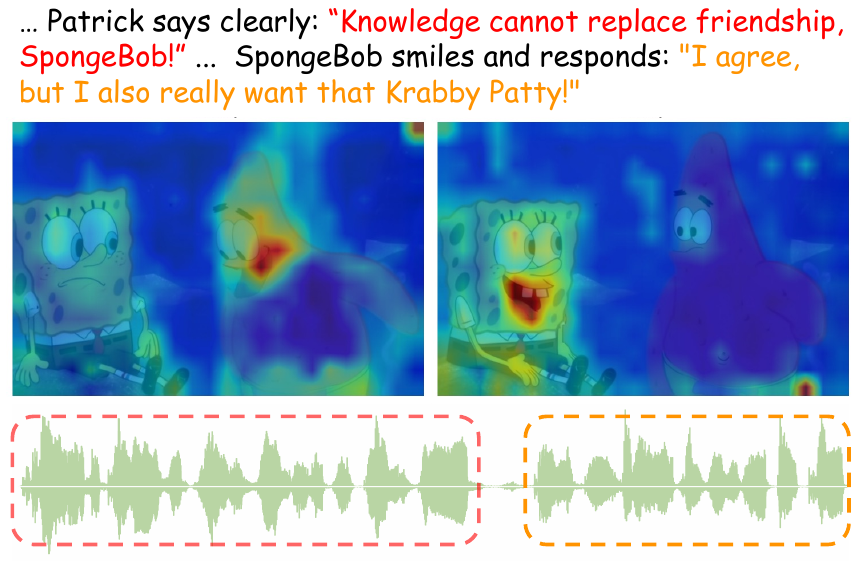}
    \caption{Visualization of V2A cross-attention maps.} 
    \label{fig:v2a_attn} 
    \vspace{-\intextsep} 
\end{wrapfigure}
In joint audio-video generation, the local visual quality of sound-emitting regions plays a decisive role in shaping the subjective perceptual experience. However, uniform updates overlook their unequal contributions to the overall quality. This motivates us to localize such critical regions and apply differentiated levels of importance and exploration to them. While directly incorporating an external detection module offers a straightforward solution, it is computationally expensive. Instead, we observe an intrinsic indicator by analyzing the attention maps of the cross-modal attention layers within the audio branch, as illustrated in Fig.~\ref{fig:v2a_attn}. We observe that these attention maps effectively highlight the speaking subjects and their sound-emitting regions within video frames. This strong correlation suggests that \textit{the V2A cross-attention map naturally serves as an intrinsic proxy for identifying critical regions in video frames}.

\section{Methodology}
\label{sec:method}

\begin{figure}[t]
    \centering
    \includegraphics[width=\linewidth]{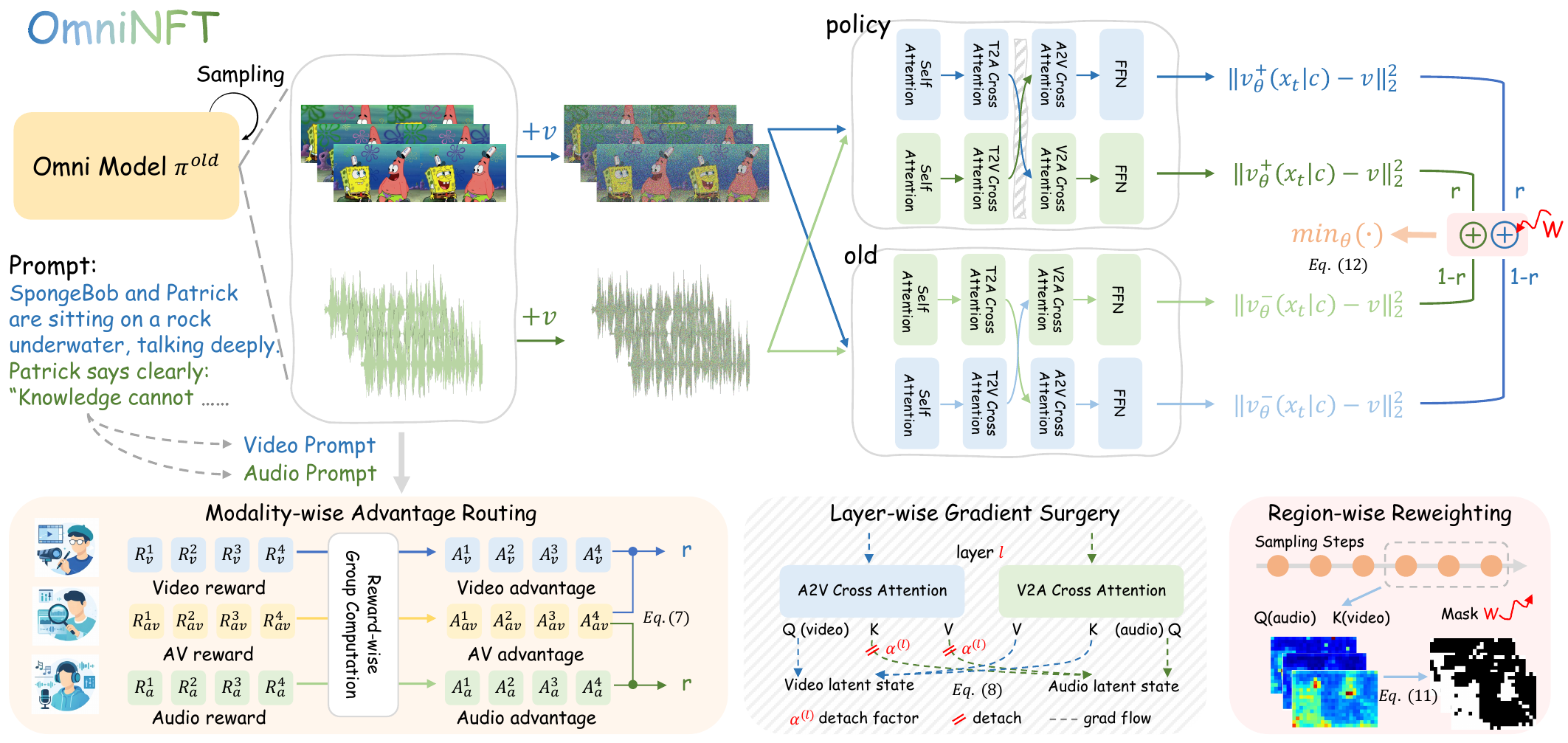}
    \caption{\textbf{Overview of OmniNFT.} Given paired video and audio prompts, the Omni model first generates joint audio-video samples. Building on these samples, OmniNFT performs three coordinated operations: (i) independent advantages derived from video, audio, and cross-modal rewards are dispatched to their corresponding branches (\textit{Modality-wise Advantage Routing}); (ii) the audio-to-video cross-attention cached from the final sampling steps is converted into a critical-region mask that reweights the RL loss (\textit{Region-wise Reweighting}); and (iii) during loss backward, the key-value gradients of A2V cross-attention in the shallow audio layers are partially detached, while all other gradient flows remain intact (\textit{Layer-wise Gradient Surgery}).}
    \label{fig:framework}
\end{figure}

Motivated by the three issues and observations discussed above in Sec.~\ref{sec:motivation}, we propose \textbf{OmniNFT}, which performs fine-grained credit assignment at three corresponding levels: modality-wise advantage routing (Sec.~\ref{sec:method_advantage}), layer-wise gradient surgery (Sec.~\ref{sec:method_surgery}), and region-wise loss reweighting (Sec.~\ref{sec:method_region}). Fig.~\ref{fig:framework} illustrates the pipeline, and Alg.~\ref{alg:omninft} summarizes the training procedure.

\subsection{Modality-wise Advantage Routing}
\label{sec:method_advantage}

\textbf{Reward-wise advantage computation.}
Instead of deriving a single advantage from all rewards, OmniNFT computes an independent advantage for each reward. During each sampling step, the model generates a group of $G$ joint audio-video pairs $\{(x_v^{(i)}, x_a^{(i)})\}_{i=1}^{G}$ conditioned on prompt $c$. For each reward function $k \in \{v, a, av\}$, we evaluate the group to obtain raw scores $\{R_k^{(i)}\}_{i=1}^{G}$ and compute a reward-wise advantage $A_k^{(i)}$ following Eq.~\ref{eq:opt_prob}, which produces three decoupled advantage sets $\{A_v^{(i)}\}_{i=1}^{G}$, $\{A_a^{(i)}\}_{i=1}^{G}$, and $\{A_{av}^{(i)}\}_{i=1}^{G}$

\textbf{Modality-decoupled advantage routing.}
With the reward-wise advantages, OmniNFT routes each advantage to the branch(es) whose output it evaluates. Specifically, the video advantage $A_v$ captures visual quality and motion, which are exclusively governed by the video branch. Analogously, the audio advantage $A_a$ reflects audio fidelity determined by the audio branch. In contrast, the synchronization advantage $A_{av}$ measures the synchronization alignment between the two modalities, and is thus \emph{broadcast} as shared supervision to both branches. The composite routed advantages are:
\begin{equation}
    \tilde{A}_v^{(i)} = A_v^{(i)} + A_{av}^{(i)}, \qquad
    \tilde{A}_a^{(i)} = A_a^{(i)} + A_{av}^{(i)}.
    \label{eq:route}
\end{equation}
This design enforces modality-specific supervision for uni-modal rewards while preserving shared cross-modal supervision, enabling more informative and less conflicting reward assignment.

\begin{algorithm}[t]
   \caption{OmniNFT: Modality-wise Omni Diffusion RL Fine-Tuning}
   \label{alg:omninft}
   \textbf{Require:} Pretrained dual-stream policy $(\vv^\text{ref}_v, \vv^\text{ref}_a)$, reward functions $\{R_v, R_a, R_\text{av}\}$, prompt dataset $\{\vc\}$, detach ratio $\alpha_s$, shallow boundary $L$, reweighting strength $w$, later denoising steps $\mathcal{T}$.\\
   \textbf{Initialize:} $\vv^\text{old}_m \leftarrow \vv^\text{ref}_m$, $\vv_{\theta_m} \leftarrow \vv^\text{ref}_m$ for $m \in \{v, a\}$, data buffer $\mathcal{D} \leftarrow \emptyset$.
   \begin{algorithmic}[1]
    \For {\text{each iteration} $i$}
        \For {\text{each sampled prompt $\vc$}} \Comment{Sampling Stage}
            \State Sample $G$ joint outputs $\{(\vx_{0,j}^{v},\, \vx_{0,j}^{a})\}_{j=1}^{G}$ from $(\vv^\text{old}_v, \vv^\text{old}_a)$.
            \State Evaluate rewards $\{R_v^{(j)}, R_a^{(j)}, R_{av}^{(j)}\}_{j=1}^{G}$ and cache V2A cross-attention maps $\{Attn^{(l,t)}\}$.
            \State {\textbf{Modality-wise advantages \& routing} (Sec.~\ref{sec:method_advantage}):} compute $A_v^{(j)}, A_a^{(j)}, A_{av}^{(j)}$, then route:
            \Statex \hspace{4.5em} $\tilde{A}_v^{(j)} \leftarrow A_v^{(j)} + A_\text{av}^{(j)}$,\quad $\tilde{A}_a^{(j)} \leftarrow A_a^{(j)} + A_\text{av}^{(j)}$.
            \State Convert $\tilde{A}_m^{(j)}$ to optimality probability $r_m^{(j)}$ via Eq.~\ref{eq:opt_prob} for $m \in \{v, a\}$.
            \State {\textbf{Region-wise weighting} (Sec.~\ref{sec:method_region}):} aggregate $\{Attn^{(j,l,t)}\}$ over $l \ge L, t \in \mathcal{T}$ to obtain $\{w^{(j)}\}$
            \State $\mathcal{D} \leftarrow \mathcal{D} \cup \{c, x_{0,j}^{v},x_{0,j}^{a},\, r_v^{(j)},\, r_a^{(j)},\, \{w^{(j)}\}\}$.
        \EndFor
        \For {\text{each mini batch} $\{c, x_0^v, x_0^a, r_v, r_a, \{w^{(j)}\}\} \in \mathcal{D}$} \Comment{Training Stage}
            \State {\textbf{Layer-wise gradient surgery} (Sec.~\ref{sec:method_surgery}):} replace audio $KV$ in A2V cross-attention block $l$ with
            \Statex \hspace{4.5em} $\tilde{K}_a^{(l)} = \alpha^{(l)} \mathrm{sg}(K_a^{(l)}) + (1-\alpha^{(l)}) K_a^{(l)}$, where $\alpha^{(l)} = \alpha_s$ if $l < L$ else $0$, same for $\tilde{V}_a^{(l)}$.
            \State Compute region-weighted video loss and standard audio loss, then update $\theta$ via Eq.~\ref{eq:total_loss}.
        \EndFor
        \State Update $\theta^\text{old} \leftarrow \eta_i \theta^\text{old} + (1-\eta_i) \theta$, and clear $\mathcal{D} \leftarrow \emptyset$. \Comment{Online Update}
    \EndFor
   \end{algorithmic}
   \textbf{Output:} $(\vv_{\theta_v}, \vv_{\theta_a})$
\end{algorithm}

\subsection{Layer-wise Gradient Surgery}
\label{sec:method_surgery}
For each Transformer block $l$, the A2V cross-attention takes its query $Q_v^{(l)}$ from the video hidden states and its key-value pair $(K_a^{(l)}, V_a^{(l)})$ from the audio hidden states. We apply a layer-wise partial detach on the these KV:
\begin{equation}
\begin{aligned}
\tilde{K}_a^{(l)} &= \alpha^{(l)}\,\mathrm{sg}(K_a^{(l)}) + (1-\alpha^{(l)})\,K_a^{(l)}, \\
\tilde{V}_a^{(l)} &= \alpha^{(l)}\,\mathrm{sg}(V_a^{(l)}) + (1-\alpha^{(l)})\,V_a^{(l)},
\end{aligned}
\end{equation}
where $\mathrm{sg}(\cdot)$ denotes the stop-gradient operator. This leaves the forward sampling unchanged but scales the backward gradient through the KV path by $(1-\alpha^{(l)})$. The detach ratio $\alpha^{(l)}$ follows a simple schedule aligned with the identified layer functionality and the above observation:
\begin{equation}
\alpha^{(l)} =
\begin{cases}
\alpha_s, & l < L \quad \text{(shallow layers)}, \\
0, & l \ge L \quad \text{(deep layers)},
\end{cases}
\end{equation}
with $L$ defaults to 10 and $\alpha_s$ is 0.1. In this way, RL gradients flow freely through the deep layers responsible for cross-modal alignment, while leakage into shallow audio layers is suppressed.

\subsection{Region-wise Reweighting}
\label{sec:method_region}

Let $Attn^{(l,t)} \in \mathbb{R}^{N_v \times N_a}$ denote the V2A cross-attention map at block $l$ and denoising step $t$, where $N_v$ and $N_a$ are the numbers of video and audio tokens. Since the V2A attention becomes semantically meaningful in the deep AV-Sync Zone and at the later denoising steps, we aggregate attention only over these informative timesteps to obtain a per-token score:
\begin{equation}
s_i = \frac{1}{|\mathcal{D}|\,|\mathcal{T}|}\sum_{l \in \mathcal{D}} \sum_{t \in \mathcal{T}} \sum_{j=1}^{N_a} Attn^{(l,t)}_{i,j}, \quad i=1,\dots,N_v,
\end{equation}
where $\mathcal{D} = \{l \mid l \ge L\}$ indexes the deep cross-modal blocks and $\mathcal{T}$ denotes the last few denoising steps. The score is then normalized and mapped into a region-wise weight:
\begin{equation}
w_i = 1 + \lambda \cdot \frac{s_i - \min_j s_j}{\max_j s_j - \min_j s_j},
\end{equation}
with $\lambda > 0$ controlling the reweighting strength.

\subsection{Overall Training Objective}
Following the DiffusionNFT formulation (Sec.~\ref{sec:Preliminary}), we convert the routed advantages into optimality probabilities $r_m$ for each branch $m \in \{v, a\}$. The region-wise weights $w_i$ are incorporated into the video branch loss to concentrate optimization capacity on perceptually critical regions, while the audio branch loss remains unchanged. Based on Eq.~\ref{eq:loss}, the total training objective becomes:
\begin{equation}
    \mathcal{L}_{all}(\theta) = \underbrace{\frac{1}{\sum_i w_i}\sum_{i=1}^{N_v} w_i \cdot \mathcal{L}_{video}^{(i)}(\theta)}_{\text{region-weighted video loss}} + \mathcal{L}_{audio}(\theta).
\label{eq:total_loss}
\end{equation}
\section{Experiments}
\label{sec:Experiments}

We first describe the experimental setup in Sec.~\ref{sec:setup}, including reward models, evaluation metrics, and implementation details. We then present the main quantitative and qualitative results in Sec.~\ref{sec:main_results}, and conduct ablation studies to analyze the contribution of each component in Sec.~\ref{sec:ablation}.

\subsection{Experimental Setup}
\label{sec:setup}
\begin{table*}[t]
\centering
\caption{Main results on JavisBench~\citep{liu2025javisdit}. Best results in \colorbox{mycolor_green}{green}, second-best \underline{underlined}.  VQ: Visual Quality, AQ: Audio Quality. ($\uparrow$: higher is better; $\downarrow$: lower is better).}
\label{tab:main_results}
\resizebox{\textwidth}{!}{%
\begin{tabular}{l l cc cccc cc cc}
\toprule
 & & \multicolumn{2}{c}{AV-Quality} & \multicolumn{4}{c}{Text-Consistency} & \multicolumn{2}{c}{AV-Consistency} & \multicolumn{2}{c}{AV-Synchrony} \\
 \cmidrule(lr){3-4} \cmidrule(lr){5-8} \cmidrule(lr){9-10} \cmidrule(lr){11-12}
Model & Size & VQ $\uparrow$ & AQ $\uparrow$ & TV-IB $\uparrow$ & TA-IB $\uparrow$ & CLIP $\uparrow$ & CLAP $\uparrow$ & AV-IB $\uparrow$ & AVHScore $\uparrow$ & JavisScore $\uparrow$ & DeSync $\downarrow$ \\
\midrule
\multicolumn{12}{l}{$-$ \textit{T2A+A2V}} \\
TempoTkn~\citep{yariv2024diverse} & 1.3B & -- & -- & 0.084 & -- & 0.205 & -- & 0.139 & 0.122 & 0.103 & 1.532 \\
TPoS~\citep{jeong2023power} & 1.0B & -- & -- & 0.201 & -- & 0.229 & -- & 0.124 & 0.129 & 0.095 & 1.493 \\
\midrule
\multicolumn{12}{l}{$-$ \textit{T2V+V2A}} \\
ReWaS~\citep{jeong2025read} & 0.6B & -- & -- & -- & 0.123 & -- & 0.280 & 0.110 & 0.104 & 0.079 & 1.071 \\
See\&Hear~\citep{xing2024seeing} & 0.4B & -- & -- & -- & 0.129 & -- & 0.263 & 0.160 & 0.143 & 0.112 & 1.099 \\
FoleyCrafter~\citep{zhang2026foleycrafter} & 1.2B & -- & -- & -- & 0.149 & -- & 0.383 & 0.193 & 0.186 & 0.151 & 0.952 \\
MMAudio~\citep{cheng2025mmaudio} & 0.1B & -- & -- & -- & 0.160 & -- & 0.407 & 0.198 & 0.182 & 0.150 & 0.849 \\
\midrule

\multicolumn{12}{l}{$-$ \textit{T2AV}} \\
JavisDiT~\citep{liu2025javisdit} & 3.1B & 1.291 & 4.478 & 0.263 & 0.143 & 0.302 & 0.391 & 0.197 & 0.179 & 0.154 & 1.039 \\

UniVerse-1~\citep{wang2025universe} & 6.4B & 1.357 & 4.839 & 0.272 & 0.111 & 0.309 & 0.245 & 0.104 & 0.098 & 0.077 & 0.929 \\

JavisDiT++~\citep{liu2026javisdit++} & 2.1B & 1.462 & 5.049 & \colorbox{mycolor_green}{0.282} & 0.164 & \colorbox{mycolor_green}{0.316} & 0.424 & 0.198 & 0.184 & 0.159 & 0.832 \\


LTX-2~\citep{hacohen2026ltx} & 19B & {2.038} & {5.197} & 0.272 & {0.170} & \underline{0.311} & {0.412} & {0.232} & \underline{0.223} & \underline{0.192} & {0.569} \\

\midrule
LTX-2+ GDPO~\citep{liu2026gdpo} & 19B & \underline{3.209} & \underline{5.523} & 0.265 & \underline{0.184} & 0.308 & \underline{0.428} & \underline{0.233} & 0.223 & 0.185 & \underline{0.412} \\

\textbf{LTX-2+OmniNFT} & 19B & \colorbox{mycolor_green}{3.326} & \colorbox{mycolor_green}{5.715} & 0.261 & \colorbox{mycolor_green}{0.189} & 0.310 & \colorbox{mycolor_green}{0.445} & \colorbox{mycolor_green}{0.262} & \colorbox{mycolor_green}{0.257} & \colorbox{mycolor_green}{0.220} & \colorbox{mycolor_green}{0.269} \\
\textcolor{gray}{\textit{Our RL $\Delta$}} & \textcolor{gray}{--} & \textcolor{gray}{+1.288} & \textcolor{gray}{+0.518} & \textcolor{gray}{-0.011} & \textcolor{gray}{+0.019} & \textcolor{gray}{-0.001} & \textcolor{gray}{+0.033} & \textcolor{gray}{+0.030} & \textcolor{gray}{+0.034} & \textcolor{gray}{+0.028} & \textcolor{gray}{-0.300} \\

\bottomrule
\end{tabular}%
}
\end{table*}



\textbf{Reward models.}
We utilize VideoAlign~\citep{liu2025improving} and HPSv3~\citep{ma2025hpsv3} scores as rewards to evaluate video quality, while Audiobox Aesthetics~\citep{tjandra2025meta} is employed as the reward for audio quality. To ensure cross-modal consistency, we adopt the CLAP~\citep{wu2023large} score as the reward for audio-text alignment, and the synchronization score (Desync)~\citep{iashin2024synchformer} as the reward for audio-visual synchronization.

\textbf{Evaluation Metrics.}
We report results across four complementary dimensions defined in JavisBench:
(i)~\textit{AV-Quality}: Visual quality (VQ) and Audio quality (AQ) from VideoAlign score for uni-modal generation fidelity. 
Furthermore, we utilize VBench~\citep{huang2024vbench} as an additional benchmark to assess the quality of the generated videos.
(ii)~\textit{Text-Consistency}: Text-Video and Text-Audio ImageBind~\citep{girdhar2023imagebind} similarity (TV-IB, TA-IB), CLIP~\citep{radford2021learning} score, and CLAP score for text-to-modal alignment;
(iii)~\textit{AV-Consistency}: Audio-Video ImageBind similarity (AV-IB) and AVHScore for semantic coherence;
(iv)~\textit{AV-synchronization}: JavisScore~\citep{liu2025javisdit} and DeSync for synchronization between audio and video.

\subsection{Main Results and Analysis}
\label{sec:main_results}

\begin{figure*}[h]
\centering
\begin{minipage}[b]{0.62\textwidth}
    \centering
    \resizebox{\textwidth}{!}{%
\begin{tabular}{l cc cc cc}
\toprule
 & \multicolumn{2}{c}{AV-Quality} & \multicolumn{2}{c}{Text-Consistency} & \multicolumn{2}{c}{AV-Consistency} \\
 \cmidrule(lr){2-3} \cmidrule(lr){4-5} \cmidrule(lr){6-7}
Setting & VQ $\uparrow$ & AQ $\uparrow$ & CLIP $\uparrow$ & CLAP $\uparrow$ & AVHScore $\uparrow$ & JavisScore $\uparrow$ \\
\midrule
\multicolumn{7}{l}{\textit{(a) Gradient Surgery Layer Selection}} \\
\midrule
Shallow layers (default) & 3.326 & 5.715 & 0.310 & 0.445 & 0.257 & 0.220 \\
Deep layers & 3.083 & 5.577 & \textbf{0.312} & 0.427 & 0.242 & 0.204 \\
\midrule
\multicolumn{7}{l}{\textit{(b) Region-wise Reweighting Loss Weight $\lambda$}} \\
\midrule
$\lambda = 1.25$  & 3.150 & 5.495 & 0.308 & 0.429 & 0.249 & 0.212 \\
$\lambda = 1.50$ (default)& \textbf{3.326} & \textbf{5.715} & 0.310 & \textbf{0.445} & {0.257} & {0.220} \\

$\lambda = 1.75$  & 2.977 & 5.714 & 0.310 & 0.438 & \textbf{0.260} & \textbf{0.223}\\
\bottomrule
\end{tabular}%
}
    \captionof{table}{Ablation study on detach layers and coefficient $\lambda$.}
    \label{tab:ablation_layer_reweight}
\end{minipage}
\hfill
\begin{minipage}[b]{0.35\textwidth}
    \centering
    \includegraphics[width=\textwidth]{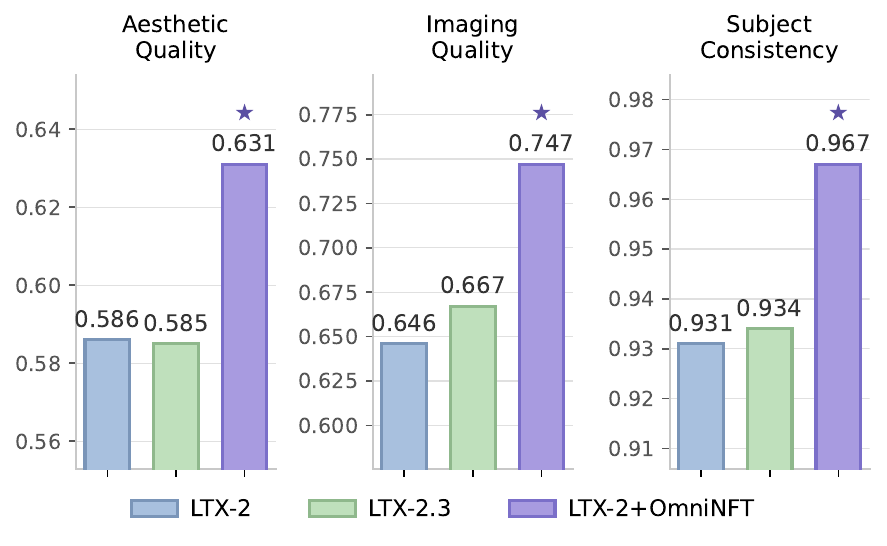}
    \captionof{figure}{Vbench Results.}
    \label{fig:vbench}
\end{minipage}
\end{figure*}

\textbf{Quantitative Analysis.}
Tab.~\ref{tab:main_results} and Fig.~\ref{fig:vbench} report results on JavisBench and VBench. 
Compared with LTX-2 (19B) and GDPO, OmniNFT achieves the best overall performance across perceptual quality, cross-modal consistency, and temporal synchronization. For perceptual quality, VQ improves from 2.038 to 3.326 (+63.2\%) and AQ from 5.197 to 5.715 (+10.0\%) over LTX-2, with notable VBench gains in imaging quality (+10.5\%). For cross-modal consistency, TA-IB surpasses both LTX-2 (+15.2\%) and the larger LTX-2.3 (22B). For synchronization, OmniNFT reduces DeSync from 0.569 to 0.269 (-52.7\%), substantially outperforming GDPO (0.412). We note that TV-IB and CLIP do not improve under either our method or GDPO, suggesting that text–video semantic alignment remains challenging.

\begin{figure}[t]
\centering
\includegraphics[width=\textwidth]{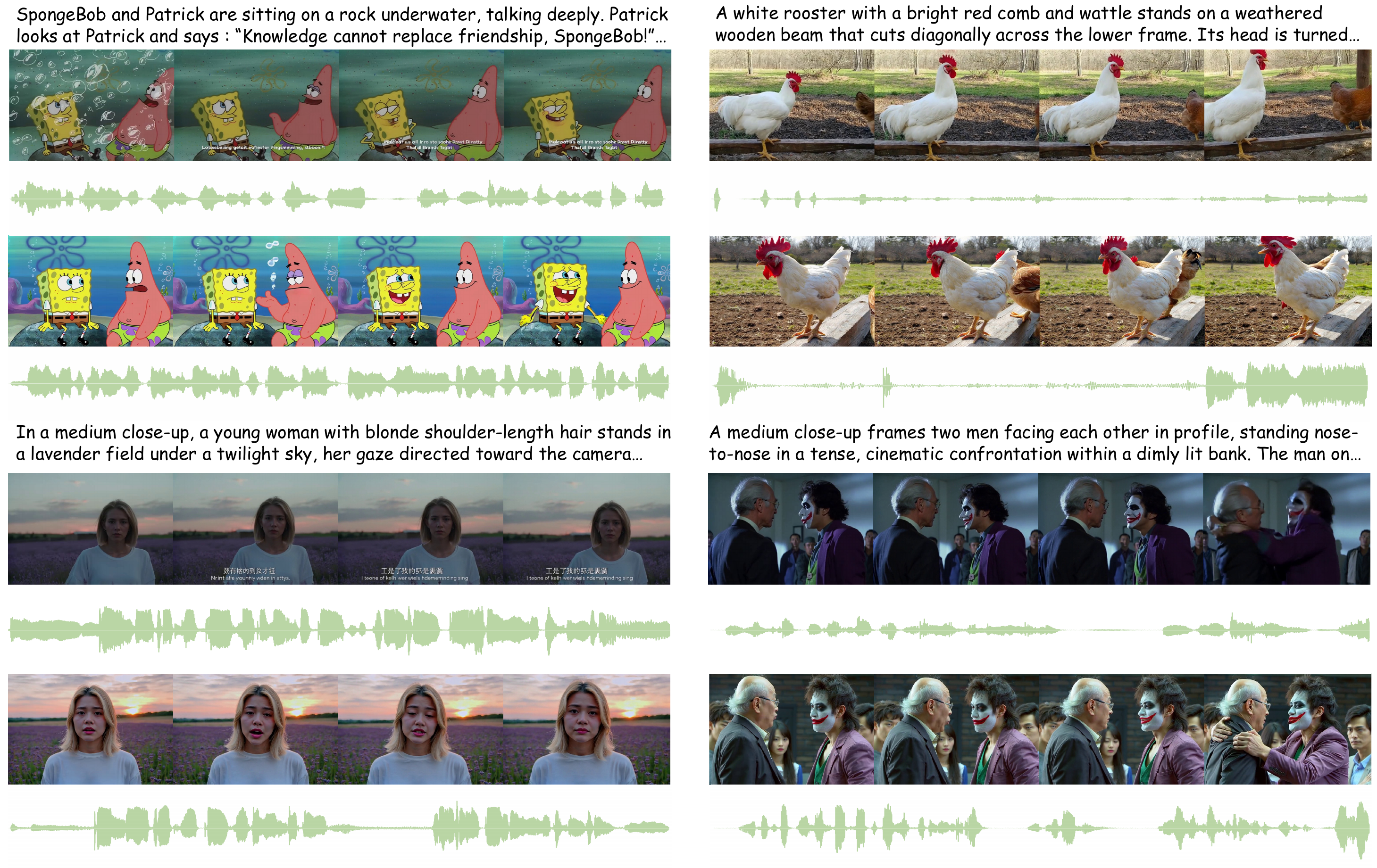}
 \caption{Qualitative examples of joint audio-video generation by OmniNFT. The four cases illustrate improvements across different aspects: enhanced visual quality, improved audio fidelity, better speech-lip synchronization, and coherent multi-speaker scenes.}
\label{fig:show_case}
\vspace{-1mm}
\end{figure}

\textbf{Qualitative Analysis.}
Fig.~\ref{fig:show_case} highlights four representative cases, each showcasing a distinct strength of OmniNFT: sharper frames and natural motion in the cartoon scene (\textit{visual quality}), richer ambient textures in the rooster case (\textit{audio fidelity}), tight waveform–lip alignment in the lavender-field close-up (\textit{speech–lip sync}), and coherent identities with alternating vocal activity in the confrontation scene (\textit{multi-speaker consistency}). These results confirm that OmniNFT produces temporally synchronized and semantically coherent audio–video content across diverse scenarios.

\subsection{Ablation Studies}
\label{sec:ablation}

\textbf{Effectiveness of Each Component.} We validate the three key designs of OmniNFT by progressively incorporating them into the vanilla RL baseline, as shown in Tab.~\ref{tab:ablation}. Decoupling advantages across modalities alleviates reward inconsistency, bringing clear improvements in cross-modal consistency and synchrony. Shielding shallow audio layers from dominant video gradients further enhances audio fidelity and text-audio alignment, while focusing synchronization-critical regions provides fine-grained credit assignment that pushes AV-consistency and synchrony to the best results. Notably, our designs introduce only a negligible overhead.

\textbf{Hyperparameter Analysis.} For gradient surgery layer $L$ selection, detaching video-to-audio gradients at shallow layers ($L<10$) consistently outperforms detaching at deep layers($L>20$) in Tab.~\ref{tab:ablation_layer_reweight}, which aligns with our observation. For the region-wise reweighting coefficient $\lambda$, a moderate value ($\lambda=1.50$) achieves the best trade-off: smaller values under-emphasize critical regions, while larger values influence visual quality.




\begin{table*}[h]
\centering
\caption{Ablation results on each design ($\uparrow$: higher is better; $\downarrow$: lower is better). Best results in \textbf{bold}.}
\label{tab:ablation}
\resizebox{\textwidth}{!}{%
\begin{tabular}{l cc cccc cc cc c}
\toprule
 & \multicolumn{2}{c}{AV-Quality} & \multicolumn{4}{c}{Text-Consistency} & \multicolumn{2}{c}{AV-Consistency} & \multicolumn{2}{c}{AV-Synchrony} & \multicolumn{1}{c}{Efficiency} \\
 \cmidrule(lr){2-3} \cmidrule(lr){4-7} \cmidrule(lr){8-9} \cmidrule(lr){10-11} \cmidrule(lr){12-12}
Model & VQ $\uparrow$ & AQ $\uparrow$ & TV-IB $\uparrow$ & TA-IB $\uparrow$ & CLIP $\uparrow$ & CLAP $\uparrow$ & AV-IB $\uparrow$ & AVHScore $\uparrow$ & JavisScore $\uparrow$ & DeSync $\downarrow$ & Time $\downarrow$ \\
\midrule
LTX-2~\citep{hacohen2026ltx} & 2.038 & 5.197 & \textbf{0.272} & 0.170 & \textbf{0.311} & 0.412 & 0.232 & 0.223 & 0.192 & 0.569 & --\\
LTX-2 + vanilla RL & 3.209 & 5.523 & 0.265 & 0.184 & 0.308 & 0.428 & 0.233 & 0.223 & 0.185 & 0.412 & \textbf{23.9h}\\
\midrule
\quad + Modality-wise advantage routing  & 3.264 & 5.399 & 0.266 & 0.186 & 0.306 & 0.430 & 0.248 & 0.240 & 0.199 & 0.322 & 23.9h\\
\quad + Layer-wise gradient surgery  & 3.246 & \textbf{5.917} & 0.264 & \textbf{0.192} & 0.311 & 0.438 & 0.252 & 0.247 & 0.209 & 0.334 & 24.1h\\
\quad + Region-wise loss reweighting (Full)  & \textbf{3.326} & 5.715 & 0.261 & 0.189 & 0.310 & \textbf{0.445} & \textbf{0.262} & \textbf{0.257} & \textbf{0.220} & \textbf{0.269} & 24.1h\\
\bottomrule
\end{tabular}%
}
\end{table*}
\section{Conclusion}
\label{sec:Conclusion}
We present OmniNFT, a modality-aware online diffusion RL framework. Through three complementary innovations: modality-wise advantage routing, layer-wise gradient surgery, and region-wise loss reweighting, OmniNFT effectively resolves advantage inconsistency, gradient imbalance, and uniform credit assignment. Extensive experiments on JavisBench and VBench demonstrate our effectiveness.

\bibliography{iclr2026_conference}
\bibliographystyle{iclr2026_conference}

\clearpage

\appendix

\end{document}